\documentclass[fleqn,10pt]{wlscirep}
\usepackage[utf8]{inputenc}
\usepackage[T1]{fontenc}
\usepackage{bm}
\usepackage[square,sort,comma,numbers]{natbib}
\title{Understanding Missingness in Time-series Electronic Health Records for Individualized Representation
}
\usepackage{comment}  

\author[1,*]{Ghadeer O. Ghosheh}
\author[2]{Jin Li}
\author[1]{Tingting Zhu}

\affil[1]{University of Oxford}

\affil[2]{Nanjing University of Information Science \& Technology (NUIST)}

\affil[*]{Correspondence: ghadeer.ghosheh@eng.ox.ac.uk}
\begin{abstract}
    With the widespread of machine learning models for healthcare applications, there is increased interest in building applications for personalized medicine. Despite the plethora of proposed research for personalized medicine, very few focus on representing missingness and learning from the missingness patterns in time-series Electronic Health Records (EHR) data. The lack of focus on missingness representation in an individualized way limits the full utilization of machine learning applications towards true personalization. In this brief communication, we highlight new insights into patterns of missingness with real-world examples and implications of missingness in EHRs. The insights in this work aim to bridge the gap between theoretical assumptions and practical observations in real-world EHRs. We hope this work will open new doors for exploring directions for better representation in predictive modelling for true personalization. 
\end{abstract}
\begin{document}

\maketitle


In the past decade, many machine learning (ML) works have been proposed for predictive modelling and representing patterns in longitudinal and time-series Electronic Health Records (EHRs). With the wealth of time-series data, many data-driven models have shown great potential in clinical research and patient care. One of the emerging areas is personalized medicine, where EHR data is used to tailor the decision-making and interventions to best fit an individual patient~\citep{abul2019personalized}. Despite the great promise shown in many applications, the current literature on personalized medicine describes methods that only focus on the observed data~\citep{chen2019deep,wang2016learning} and often overlook the "individualized" and informative nature of unobserved or "missing" values in the data. In clinical practice, EHRs tend to be highly missing and irregularly sampled~\citep{madden2016missing}, where missingness can reach up to 80\% in the intensive care unit settings (see MIMIC~\citep{johnson2016mimic} and eICU~\citep{pollard2018eicu}), despite patients being monitored continuously. Therefore, there is a pressing need to learn proper patient representation and analysis of data missingness to improve ML models for better patient outcomes. 

While there are existing works for handling missingness in time-series EHRs, most of them assumed missingness was under the context of missing at completely random (MCAR)~\citep{yoon2018gain,cao2018brits,fortuin2020gp}, which may not be true in reality. In the case of MCAR, it is assumed that the missingness in the data is related to neither the observed nor unobserved data~\citep{mack2022types}. Such a strong assumption disregards any information related to the frequency and the reasons for recording, which might be related to the patient's health status. 
For example, missing values in time-series EHRs can occur due to various reasons such as recording errors and machine failure, irregular sampling and inconsistent medical visits~\citep{kreindler2006effects}, or even high-cost and dangerous to acquire information such as invasive or radiology procedures~\citep{bulas2013benefits,kim2017dangers}. In addition, the measurement frequency could also be related to factors such as patient severity and  deterioration~\citep{agniel2018biases,ghosheh2023survey}, lack of medical need~\citep{duff2019frequency}, bias and quality of care~\citep{weber2021gender,wells2013strategies}, all of which vary from one patient to another. This highlights the importance of representing the individualized missingness in personalized models.  Recent ML models for healthcare have shown that the utilization of binary missingness masks to indicate the presence of observations as input features can result in equivalent performance to those built on the observed values of clinical features~\citep{sharafoddini2019new,che2018recurrent}, indicating the predictive value of missingness patterns. However, despite the utility and predictiveness of the binary missingness masks, they are simplistic and lack indications of the patterns and frequencies of individual-level missingness for each patient.

In this brief communication, we highlight new insights into patterns of missingness in EHRs, which require a new perspective on representing missingness beyond binary masks.
In time-series EHRs collected in hospitals, the measurement frequency of a variable over time often indicates the patient's underlying state.  We refer to the missingness of observations of a variable across a patient record as feature-wise missingness. Precisely, some lab values are only measured for severe patients to differentiate clinical complications, such that a single measurement of a feature is enough to diagnose the patient.  For example, cardiac Troponins, sensitive biomarkers of myocardial injury, are not ordered for every patient in the ICU but are for those suspected of having a cardiac-related diagnosis or complication ~\citep{daubert2010utility}. Hence, a missing recording of cardiac Troponins can indicate that the patient is not suspected of developing cardiac complication.
In such cases, representing the feature-wise missingness over timesteps via a binary mask would miss on representing patterns in an individualized way. 

\begin{figure}
    \centering
    \includegraphics[width = 0.85\linewidth]{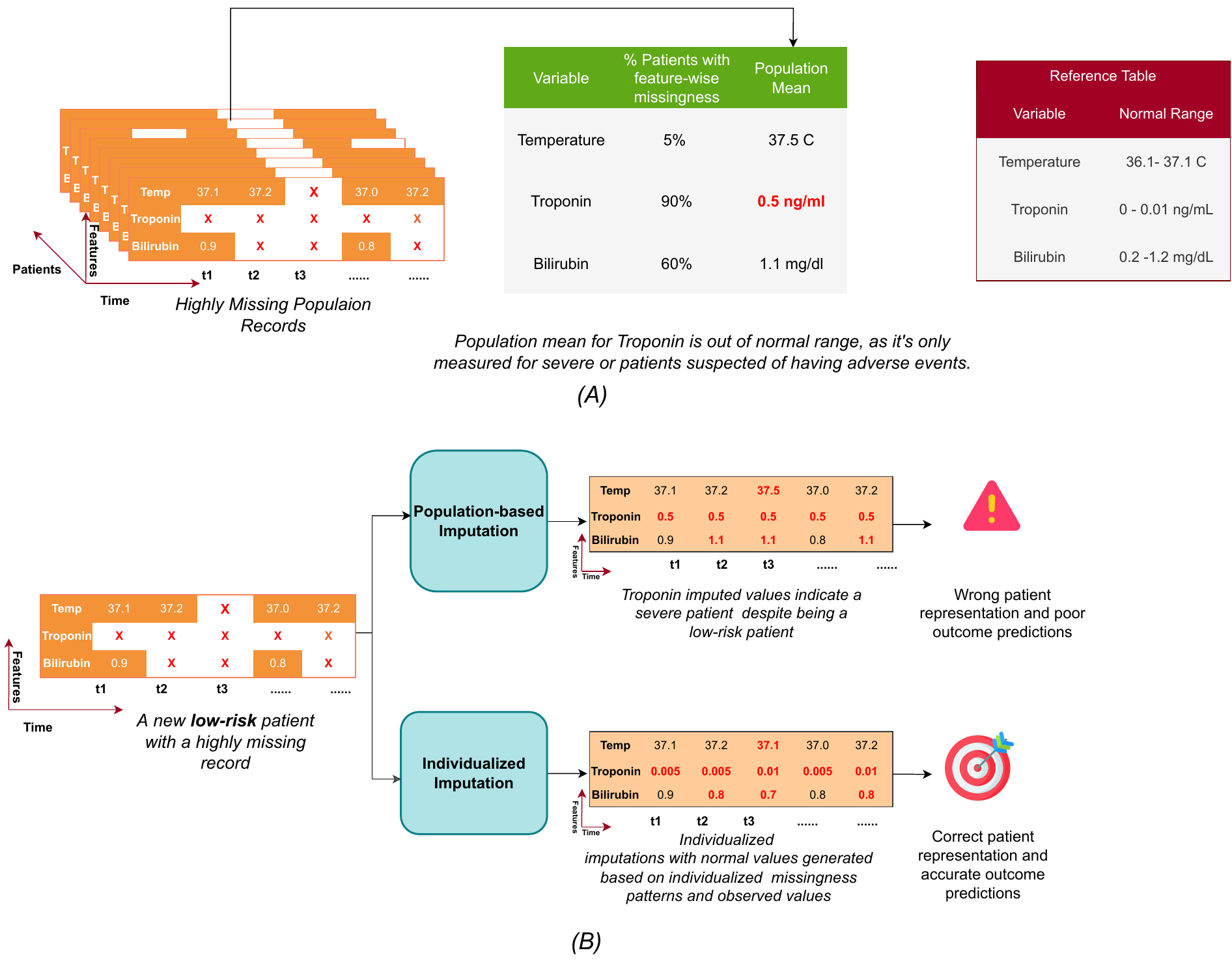}
    \caption{An example showing the impact of different imputation approaches for time-series EHRs with feature-wise missingness. In (A), highly missing EHRs are shown with population-level statistics and reference tables. In (B), the impact of individualized imputation compared to population-based imputation is shown to affect patient representation and predictive modelling, respectively.}
    \label{fig:enter-label}
\end{figure}
Understanding the difference between general missingness in a patient record and feature-wise missingness for some variables is crucial to recognizing the implications of imputation. For example, filling the missing cardiac troponin values for healthy patients by the population mean will result in an imputation value above the healthy range since all the available measurements are for patients suspected of having MI. Such imputations would result in a wrong patient representation and falsely indicate patient severity for healthy patients, resulting in a flawed data distribution. In Figure 1, we showcase an example where population-based imputation methods fail to impute the data to represent a patient on a personalized level. Due to the limitations of such simple imputation techniques in handling individualized patterns in missingness and introducing bias in data distribution, we believe new approaches to representing missingness in predictive models can help generate high-quality imputations and build high-performance and robust predictive models.  
Compared to traditional population-level imputations, conditional and generative imputation models~\citep{yoon2018gain,fortuin2020gp, du2023saits} generate realistic imputations by learning the underlying patient population distribution in the observed data.
However, they tend to lack learning from missingness patterns in the data, limiting the models to learning from a highly predictive data component, making the models better suited for personalized medicine. In recent work in Individualized GeNeration of Imputations in Time-series Electronic health records (IGNITE)~\citep{ghosheh2024ignite}, authors present a deep generative model that generates imputations based on personalized patterns in observed and missing data for each patient. Introducing a new frequency-based individualized Missingness Mask (IMM), where the binary mask is transformed into an individual-level mask reflective of the frequency of measurement of features across time, allows the model to learn from personalized missingness patterns. 
We believe insights related to feature-wise missingness and other real-world implications of missingness would open new doors for exploring directions for better representation in predictive modelling. To this end, the imputation and learning missingness representations in the EHRs are beyond a preprocessing step; making the field incomplete without proper handling and representation of missingness in the data.

\bibliography{sample}
\end{document}